%% file: 2020-ablett-corrective-techreport.tex
\documentclass{stars-techreport}

\usepackage[noadjust]{cite}
\usepackage{xparse} 
\usepackage{amsmath} 
\usepackage{amssymb}
\usepackage{amsfonts} 
\usepackage{bbm}
\usepackage{color}
\usepackage{mathtools}
\usepackage{multirow}
\usepackage{graphicx}
\usepackage{subcaption}
\usepackage{wrapfig}
\usepackage{calc}
\usepackage{balance}

\usepackage{booktabs}

\usepackage{url}

\usepackage{algpseudocode}
\usepackage{algorithm}

\usepackage{hyperref}
\usepackage[capitalize]{cleveref}

\input{notation-shortcuts}

\STARStitle{Fighting Failures with FIRE: Failure Identification to\\ Reduce Expert Burden in Intervention-Based Learning}
\STARSdate{August 10, 2020} %
\STARSauthor{Trevor Ablett, Filip Mari\'{c}, and Jonathan Kelly}  %
\STARSidentifier{STARS-2020-001}  %
\STARSmajorrev{1}  %
\STARSminorrev{03} %

\begin{document}
\maketitle
\thispagestyle{empty}

\begin{abstract}
	Supervised imitation learning, also known as behavioral cloning, suffers from distribution drift leading to failures during policy execution. One approach to mitigate this issue is to allow an expert to correct the agent's actions during task execution, based on the expert's determination that the agent has reached a `point of no return.' The agent's policy is then retrained using this new corrective data. This approach alone can enable high-performance agents to be learned, but at a substantial cost: the expert must vigilantly observe execution until the policy reaches a specified level of success, and even at that point, there is no guarantee that the policy will always succeed. To address these limitations, we present \Method \space (\MethodLong), a system that can predict when a running policy will fail, halt its execution, and request a correction from the expert. Unlike existing approaches that learn only from expert data, our approach learns from both expert and non-expert data, akin to adversarial learning. We demonstrate experimentally for a series of challenging manipulation tasks that our method is able to recognize state-action pairs that lead to failures. This permits seamless integration into an intervention-based learning system, where we show an order-of-magnitude gain in sample efficiency compared with a state-of-the-art inverse reinforcement learning method and dramatically improved performance over an equivalent amount of data learned with behavioral cloning.
\end{abstract}

\section{Introduction}

Imitation learning has proven to be an effective approach to overcome many of the limitations of reinforcement learning for robotic agents: it can significantly reduce the sample complexity of pure reinforcement learning \cite{sunDeeplyAggreVaTeDDifferentiable2017} and it can eliminate the need for hand-designed rewards, enabling agents to operate directly from observations without access to the state information often required by reward functions. In some cases, even simple supervised imitation learning (also known as behavioral cloning), in which a model is trained directly on a dataset of expert state-action pairs, can achieve impressive results \cite{zhangDeepImitationLearning2018, pomerleauALVINNAutonomousLand1989, bojarskiEndEndLearning2016, giustiMachineLearningApproach2016}. Unfortunately, policies learned through behavioral cloning will often fail \cite{rossReductionImitationLearning2011} unless they are initially presented with an abundance of data that can be physically very costly to acquire, especially from human experts. Additionally, policies learned with this paradigm may work in many cases, but may also fail in surprising ways \cite{zhangDeepImitationLearning2018, pomerleauALVINNAutonomousLand1989}.

These two issues, which we will refer to as covariate shift and failure recognition respectively, have both received significant attention in the literature. Covariate shift can be resolved by having an expert relabel every state that an agent encountered during execution in a process known as dataset aggregation, or DAgger \cite{rossReductionImitationLearning2011}. This solution, and other related approaches, is very effective when a programmatic expert is available to autonomously relabel states \cite{panAgileAutonomousDriving2018}, but for many robotics tasks the only expert available is a human being. In such cases, except for trivially simple action spaces, it can be difficult or even impossible for a human to provide an adequate label: given the offline setting, the human has no feedback on whether the magnitudes of their action labels are correct. Existing work has shown that policies learned with offline human labels in a DAgger framework tend to be unstable \cite{kellyHGDAggerInteractiveImitation2019, rossLearningMonocularReactive2013}. Furthermore, in its standard form and with only a human expert available, DAgger requires a learned (or \textit{novice}) policy to execute without intervention, meaning that the policy will reach many failure states or irrelevant states before a desirable policy is learned---this is problematic, and potentially catastrophic, for real robots interacting with the physical world.

\begin{figure}
	\centering
   	\includegraphics[width=0.75\textwidth]{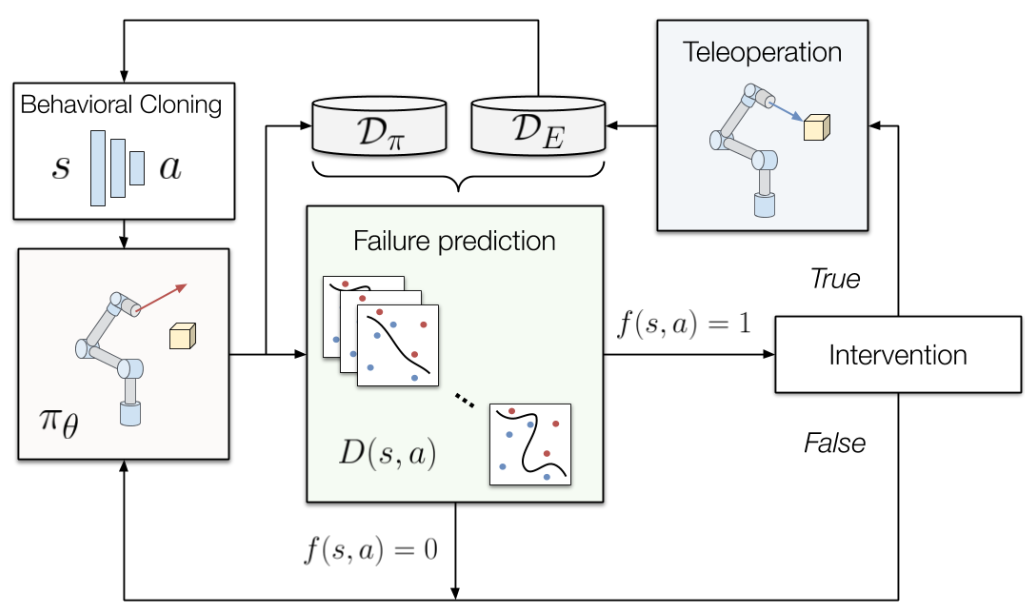}
   	\caption{An overview of our proposed technique for learning policies through intervention-based learning with predicted failures. An initial policy is learned with a handful of demonstrations through behavioral cloning (top left), which the agent then executes. A failure predictor that uses a discriminator to classify expert and non-expert data predicts if and when a failure is likely to occur (center). When a failure is predicted ($f(s,a) = 1$), execution of the policy halts, and a human observer intervenes by either agreeing (True) or disagreeing (False) with the prediction. If they agree with the prediction, they teleoperate the robot to move it into a new state where they believe the existing policy can complete the task (top right). Periodically, the policy is retrained on both existing and new expert data (top left).}
   	\label{fig:view_agnostic_results}
\end{figure}

Much of the existing work on failure recognition and prediction in imitation learning reduces to out-of-distribution detection \cite{mendaEnsembleDAggerBayesianApproach2019,laskeySHIVReducingSupervisor2016, kellyHGDAggerInteractiveImitation2019, kimMaximumMeanDiscrepancy2013, cuiUncertaintyAwareDataAggregation2019}, that is, attempting to recognize state-action pairs that do not resemble those in the expert dataset. These methods require setting a threshold for determining an allowable distance from the expert distribution, which can be nontrivial. A different method for predicting failures involves training a safety policy that attempts to predict whether an individual policy action deviates from the expert \cite{zhangQueryEfficientImitationLearning2017}. While this method is able empirically to reduce the expert relabelling burden in DAgger, it suffers from the limitation (noted in \cite{mendaEnsembleDAggerBayesianApproach2019}) that it does not consider the policy's epistemic uncertainty---a single threshold may be too conservative in certain states, but too permissive in uncertain states.

We present \Method, a technique for attempting to resolve covariate shift and failure recognition simultaneously while learning a policy for completing a particular task. Compared to DAgger-based approaches for solving covariate shift, a human expert directly corrects the policy \textit{during} execution and receives feedback on the effects of their actions immediately, making it possible for them to provide higher-quality labels than in DAgger with offline human labelling. Recent work has shown an online intervention-based approach can effectively learn a policy in self-driving car \cite{kellyHGDAggerInteractiveImitation2019} and drone landing \cite{goecksEfficientlyCombiningHuman2019} domains.

Intervention-based learning suffers from high expert burden: an expert must be vigilant throughout execution of the novice policy for an indefinite number of trials. To address this, we incorporate failure prediction into our framework, so that when our system predicts a failure, motion can be halted and the human expert can add corrective data. Of course, this approach only works for tasks that are quasistatic, with relatively low speeds throughout the motion, but manipulation tasks that the community is interested in fall into this category. Our approach builds on adversarial inverse reinforcement learning methods that learn a discriminator between expert and non-expert data which doubles as a reward function \cite{hoGenerativeAdversarialImitation2016, fuLearningRobustRewards2018}. We use the learned discriminator and a threshold that is modified during policy execution based on human feedback to predict failures, effectively reducing the burden on the expert. Specifically, we make the following novel contributions in this note:

\begin{enumerate}
	\item We present a new method (\Method) for predicting failures based on learning a discriminator and combining its output with a threshold value that is adjusted automatically based on human preferences during execution.
	\item We show that an \Method, an online-intervention-based approach with dataset aggregation, can learn effective policies in a variety of robotic manipulation domains. This learning occurs with significantly better sample complexity than a state-of-the-art inverse reinforcement learning-based technique and with greater stability than behavioral cloning.
	\item We compare our failure prediction results with an existing method for identifying unsafe states.
\end{enumerate}

\section{Background and Related Work}

We formulate the agent learning problem as a Markov Decision Process (MDP), as is common for learning-based solutions. Our goal is to learn a policy $\pi: S \rightarrow A$ for the environment states $s \in S$ and actions $a \in A$ that has a minimal divergence from an \textit{expert} policy $\pi_E$. We specifically note that $\pi_E$ may not be an \textit{optimal} policy for an environment, in that it is approximated by a sample of human-generated expert demonstrations $\mathcal{D}_E := \{\tau_1, \dots, \tau_M\}, \tau_m := \{(s_1, a_1), \dots, (s_T, a_T) \}$, where our horizon length is $T$. These demonstrations do not necessarily achieve maximal reward under the environment constraints due to varying degrees of sub-optimality in each individual expert action. 

\subsection{Behavioral Cloning and DAgger}

A simple, but often very effective, approach for learning an agent given expert demonstrations is to simply treat $\mathcal{D}_E$ as a dataset and use supervised learning, often referred to as \textit{behavioral cloning (BC)} \cite{pomerleauALVINNAutonomousLand1989}. We can model $\pi_\theta(a\,|\,s)$ using any function approximator and then learn its parameters $\theta$ by maximizing the likelihood
\begin{align*}
	\max_{\theta} \sum_{(s,a) \in \DE} \log \pith(a\,|\,s).
\end{align*}
Typically, in BC, we consider the deterministic alternative $\pi(s)$ to $\pi(a\,|\,s)$ and we modify our objective to minimize squared error
\begin{align} \label{eq:bc_mse}
	\min_{\theta} \sum_{(s,a) \in \DE} \Norm{\pi_\theta(s) - a }^2_2.
\end{align}

BC on its own will typically fail, in at least some cases, because the distribution of states encountered when executing $\pi_\theta$ does not necessarily match the distribution found in $\mathcal{D}_E$. This problem is resolved by DAgger \cite{rossReductionImitationLearning2011}, in which a BC-trained policy iteratively executes and relabels the states in the recorded dataset $\mathcal{D_\pi}$ with their corresponding expert actions. The new dataset is appended to the existing dataset, and the policy is re-trained. DAgger has the appealing property that, given a few assumptions, the amount of expert data required to achieve a policy with the same performance as $\pi_E$ is $O\left(T \log(T) \right)$. As stated, this strategy can be non-trivial to implement when the expert is a human. It is much easier for a human to provide labels by actually directly controlling the agent, allowing the $(s,a)$ pairs to be recorded naturally as they occur.

\subsection{Online Intervention-Based Learning}

To attempt to leverage the convergence properties of DAgger while allowing a human to provide high-quality labels, a human observer can predict that a policy will fail, intervene, return the agent to a state closer to the distribution $\mathcal{D}_E$, and then return control to $\pi_\theta$. Strictly speaking, this approach is not the same as DAgger, since only the predicted failure states are relabelled with their corresponding expert actions, and the expert subsequently provides data to recover. This ensures that $\pith$ is not allowed to execute until failure, and does not require the human to provide labels offline. 

This approach has been empirically shown to have varying degrees of success \cite{goecksEfficientlyCombiningHuman2019, cabiScalingDatadrivenRobotics2020, kellyHGDAggerInteractiveImitation2019}. An obvious limitation of this approach is that it requires a vigilant human expert to perpetually watch the agent execute its policy, and be able to react quickly enough to take over control from the agent at appropriate times. This can be difficult due to an individual's inherent delayed reaction time. For the general goal of finding an effective policy, the process may be exhausting for a human, since there is no guarantee of how many corrections the individual will have to provide before the policy reaches satisfactory performance.

\subsection{Safe Intervention-Based Learning}

Our work is most similar to \cite{kellyHGDAggerInteractiveImitation2019}, in which the authors use expert feedback to find a threshold for determining when a policy may take an unsafe action. The authors of \cite{kellyHGDAggerInteractiveImitation2019} formulate their policy as an ensemble, following \cite{mendaEnsembleDAggerBayesianApproach2019}, which provides estimates of epistemic uncertainty for the policy actions. The mean of the measured uncertainty of state-action pairs before human corrections is employed as the threshold for determining what degree of policy uncertainty is considered safe. In other words, policy uncertainty is considered to be inversely proportional to safety. The authors claim that their uncertainty method could be used to notify an expert when the agent has reached an unsafe state, but they do not actually incorporate this ability into their system.

Our method for predicting failures is similar to but distinct from another popular method for reducing expert burden \cite{zhangQueryEfficientImitationLearning2017}, in which a ``safety" policy is learned that outputs whether an action is safe or not based on the mean squared error between it and an expert action. In \cite{zhangQueryEfficientImitationLearning2017}, an oracle policy must still be available to be arbitrarily queried to update the safety policy, however, as is the case in DAgger. We do not assume to have access to an oracle policy, making it impossible to use \cite{zhangQueryEfficientImitationLearning2017} in our case.

\subsection{Inverse Reinforcement Learning (IRL)}

An alternative approach to imitation learning involves trying to learn a reward function $r(s,a) : S \times A \rightarrow \Real$ that is maximized by $\pi_E$. State-of-the-art approaches to this problem are based on a generative adversarial network (GAN) \cite{goodfellowGenerativeAdversarialNets2014} framework where $\pit$ is treated as a generator, and a discriminator $D_w : S \times A \rightarrow [0, 1]$, doubling as a reward function, is learned to discriminate between state-action pairs generated by $\piE$ and $\pit$ \cite{hoGenerativeAdversarialImitation2016, fuLearningRobustRewards2018, kostrikovDiscriminatorActorCriticAddressingSample2019}. In these methods, the discriminator is trained using a cross entropy loss
\begin{equation} \begin{aligned} \label{eq:cross_entropy}
	\max_{\theta} \min_w 
	& \sum_{(s,a) \in D_{\pi_{\theta}}} \left[ \log(D_w(s,a)) \right] +
	  \sum_{(s,a) \in \DE} \left[ \log(1 - D_w(s,a)) \right],
\end{aligned} \end{equation}
which is directly minimized when training $D_w$, and indirectly maximized by taking an optimization step based on policy gradients (e.g. \cite{schulmanTrustRegionPolicy2015, schulmanProximalPolicyOptimization2017}) that maximizes trajectory reward 
\begin{align*}
	\ExpectationSamp{(s_t,a_t) \sim \pit}{\sum_{t=1}^{T} D_w(s_t, a_t)}.
\end{align*}

Off-policy versions of these methods have been developed to attempt to reduce their high sample complexity \cite{kostrikovDiscriminatorActorCriticAddressingSample2019, sasakiSampleEfficientImitation2019}. Rather than only using the samples from the latest policy $\pit$, $D_w$ is learned by sampling from a replay buffer $\mathcal{R}$ of all previous $(s,a)$ pairs encountered during all iterations of $\pit$.
These methods have been shown to reduce the sample complexity of their on-policy counterparts considerably, but still tend to require a number of environment interactions that may be impractical on a real robot. For example, the authors of \cite{kostrikovDiscriminatorActorCriticAddressingSample2019} show that in a (simulated) robotic pushing task, over 400k environment interactions were required before a stable policy was achieved, and that policy's performance was still considerably lower than that of $\piE$. Moreover, like any other technique with a reinforcement learning component, these policies will fail numerous times before achieving adequate performance. In \cite{hoGenerativeAdversarialImitation2016}, the authors hypothesize that initializing policies with BC will lower the sample complexity, but as pointed out in \cite{sasakiSampleEfficientImitation2019, jenaAugmentingGAILBC2020}, this does not seem to be the case and requires further investigation. Our method for predicting failures is inspired by the discriminator trained in off-policy adversarial IRL methods, but we do not use the discriminator directly to improve our policy.

\section{\Method \space (\MethodLong)}
\label{sec:learning_approach}

The follwing sections describe our algorithm and its individual components.

\subsection{Predicting Failures with a Discriminator}
Our method for predicting failures is based on a principle that assumes that we have access to $D^*$, an optimal discriminator. It is known that $D^*(s, \piE(s)) = 0.5 \: \forall s\in S$ because a discriminator trained to optimality outputs 0.5 for all values in the desired distribution, and for any other policy $\pith$, $D^*(s, \pith(s)) < 0.5$ \cite{goodfellowGenerativeAdversarialNets2014}, assuming that a policy (generator) was simultaneously trained to optimality. If the policy did not reach optimality, but we somehow knew that the discriminator was optimal, $D^*(s, \piE(s)) \ge 0.5 \: \forall s\in S$. Therefore, if we had $D^*$, we could use the following rule to predict failures with a failure predictor $f : S \times A \rightarrow \{0, 1\}$, where $f(s, a) = 1$ indicates a failure:
\begin{align} \label{eq:fail_pred_ideal}
	f(s_t, a_t) = 
	\begin{cases}
		D^*(s_t, a_t) \ge 0.5 &\quad 0, \\
		\text{else} &\quad 1.
	\end{cases}
\end{align}

Of course, we will never have $D^*$, but merely the approximation of it, $D$. In this case, we can still use the same principle to guide our design for $f$; we choose to add a new hyper-parameter to account for approximation error:
\begin{align} \label{eq:fail_pred}
	f(s_t, a_t) = 
	\begin{cases}
		\{ D(s_k, a_k) < 0.5 \}_{k=t-\beta,\dots,t} &\quad 1, \\
		\text{else} &\quad 0.
	\end{cases}
\end{align}

We can set $\beta \in \mathbb{N}$ to allow us to predict a failure based on having $\beta$ consecutive estimates from $D$ below 0.5. The larger the value used for $\beta$, the less likely the system is to predict a failure. %

Naturally, adding free parameters to any method is undesirable, and an ideal value for $\beta$ would vary based on the environment, the tolerance for risk, the current accuracy of $D$, and the current performance of $\pi_{\theta}$. In our case, similar to \cite{kellyHGDAggerInteractiveImitation2019}, we use human feedback during execution to indirectly adjust the parameter, accounting for all of these factors simultaneously without forcing the human expert to directly supply $\beta$ themselves. As is common in adversarial imitation learning approaches, we train our discriminator after every episode with a cross-entropy loss (\cref{eq:cross_entropy}) by sampling random batches of data from $\Dpi$ and $\DE$ with a number of gradient steps equivalent to the episode length.

\begin{algorithm}[]
    \caption{Intervention-Based Learning with \Method}
    \small
    \label{alg:main_alg}
    \begin{algorithmic}[1]
    \State \textbf{Input}: Expert dataset of state-action pairs $\DE$, $\beta$, $\delta_{\text{fn}}, \delta_{\text{fp}}$, $d$, human observer $H$ \\
    Initialize $D_w$ randomly. \\
    Initialize policy $\pit \leftarrow$ \textit{BehaviorClone($\pit$, $\DE$)} \\
    Initialize policy data $\Dpi$ with initial episodes
    \While {task performance is unsatisfactory}
    	\For {$t = 1, \dots, T$}
    		\State Failure prediction $f(s_t, a_t) \leftarrow$ \cref{eq:fail_pred}    		
    		\If {$f(s_t, a_t) = 1$ (predicted failure)}
    			\State stop execution; query human: will agent fail?
    			\If {yes (\textit{true positive})}
    				\State \texttt{human\_control} $\leftarrow 1$
    			\Else {\space (\textit{false positive})}
    				\State $\beta \leftarrow \beta + \delta_{\text{fp}}$
    			\EndIf
    		\ElsIf {human predicts failure (\textit{false negative})}
    			\State \texttt{human\_control} $\leftarrow 1$
    			\State $\beta \leftarrow \beta - \delta_{\text{fn}}$
    		\EndIf
    		
    		\If {\texttt{human\_control}}
    			\State $a_t \leftarrow H(s_t)$
    			\State $\DE \leftarrow \DE \cup (s_t, a_t, s_{t+1})$
    		\Else
    			\State $a_t \leftarrow \pi(s_t)$
    			\State $\Dpi \leftarrow \Dpi \cup (s_t, a_t, s_{t+1})$
    		\EndIf
    	\EndFor
    	\State Update $D_w$ with \cref{eq:cross_entropy} for $T$ steps
   		\State Update $\pith$ with $T$ sampled batches from $\DE$ and $T$ gradient steps using \cref{eq:bc_mse}
   		\If {$|\DE|\mod d = 0$}
   			\State $\pit \leftarrow$ \textit{BehaviorClone($\pit$, $\DE$)}
   		\EndIf
    \EndWhile
    \Function{BehaviorClone}{$\pi$, $\mathcal{D}$}
    	\State Train $\pi$ with $\mathcal{D}$ and \cref{eq:bc_mse} by minimizing validation error
    \EndFunction
    \end{algorithmic}
\end{algorithm}

\subsection{Automatic Adjustment of  $\beta$}
Our full algorithm is shown below (\cref{alg:main_alg}). Adjustments to $\beta$ are made based on three possible responses to the failure prediction system:

\begin{enumerate}
	\item \textit{Correct prediction (true positive)}: The human expert agrees with the prediction, meaning that they agree that the agent will fail if it continues to execute, in which case $\beta$ remains the same.
	\item \textit{Incorrect prediction (false positive)}: The human expert disagrees with the prediction, meaning that the expert believes the agent will not fail if it continues to execute. $\beta$ is increased by $\delta_{\text{fp}} \in \mathbb{N}^+$ to make the system more permissive.
	\item \textit{Missed prediction (false negative)}: The human expert manually stops the system when they believe a failure should have been predicted, but was not. $\beta$ is decreased by $\delta_{\text{fn}} \in \mathbb{N}^+$ to make the system more conservative.
\end{enumerate}

The modification hyperparameters $\delta_{\text{fp}}$ and $\delta_{\text{fn}}$ are easier to manually tune than $\beta$ directly, and can be set based on an expert's own tolerance for false positives and false negatives (i.e., higher settings indicate a lower tolerance). While there are numerous other possibilities for determining how to modify $\beta$, we found that this simple scheme performed adequately. %

\subsection{Other Algorithm Considerations} 
\label{sec:other_alg_cons}

As detailed in \cref{alg:main_alg}, and as is done in \cite{goecksEfficientlyCombiningHuman2019}, although our policy is constantly retraining on samples from the expert, we retrain on the full batch of expert samples whenever $d$ new expert samples are added by minimizing the loss of a random subset of validation expert data.

\begin{figure*}[h!]
   	\includegraphics[width=.98\textwidth, height=4cm]{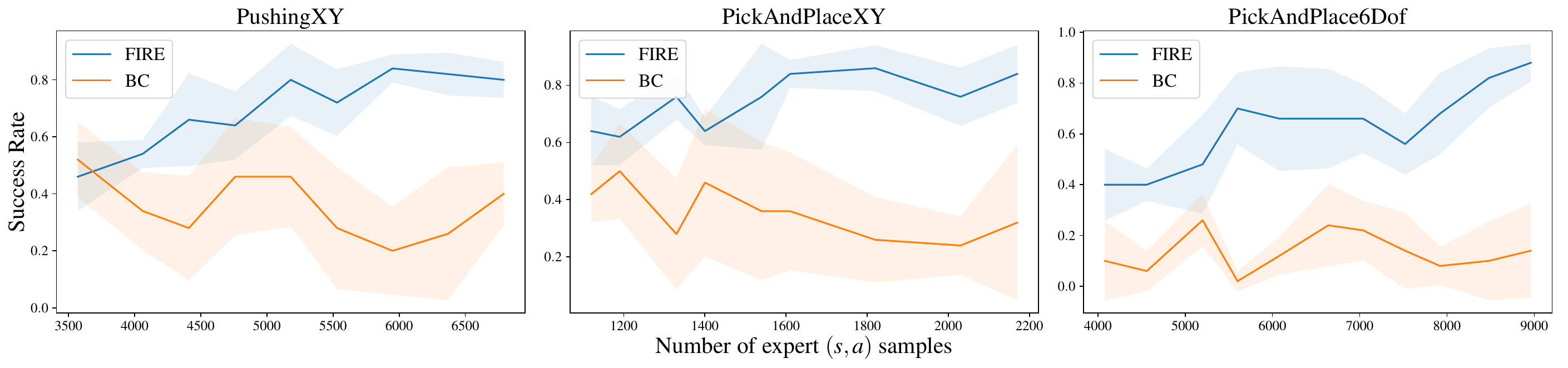}
   	\caption{Performance of \Method \space and BC in each of our test environments, compared with total number of expert $(s,a)$ samples. Due to the suboptimality of human demonstrations, behavioral cloning has high variance, and adding more data does not necessarily improve performance. As well, the corrective expert data we collect throughout the execution of \Method \space ensures that we address the distribution drift that occurs during standard BC, without requiring the collection of costly offline labels, as in \cite{rossReductionImitationLearning2011}.}
   	\label{fig:data_efficiency}
\end{figure*}

Since in our online intervention setting we have access to a human expert, we assume that we receive a binary sparse reward signal indicating whether each episode ends successfully. In practice, this is used to add $(s_T, a_T)$ of a successful episode to $\DE$, regardless of whether $a_T$ was an expert action or not, to ensure that $D$ always gives high value to successful states. We argue that this is a fair assumption, since most finite-horizon benchmark environments will output a ``done" signal when a goal is reached.

\begin{figure}[h!]
   	\includegraphics[width=.98\columnwidth]{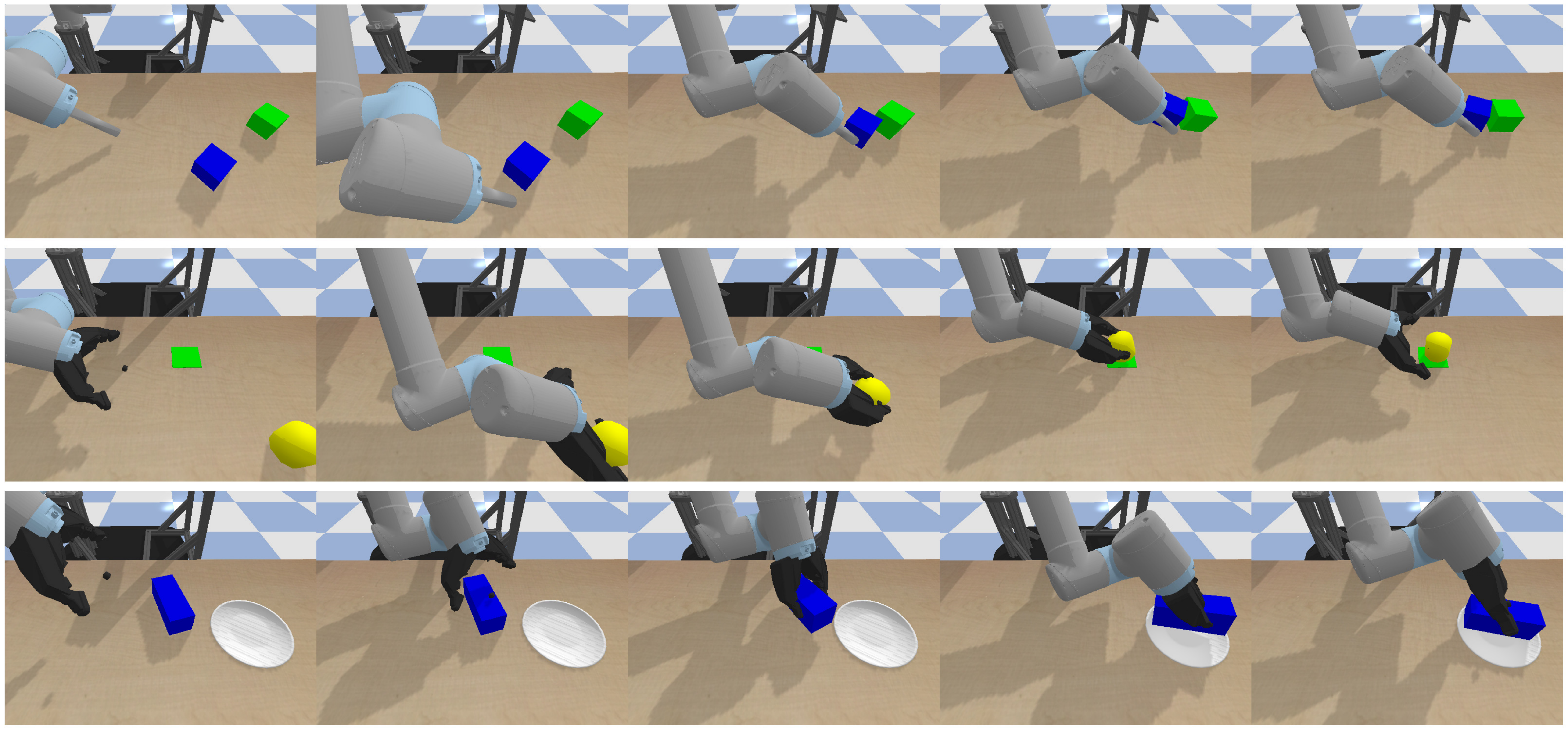}
   	\caption{Examples of successful episodes in each of our environments, showing frames from $t=\{1, .25T, .5T, .75T, T\}$.}
   	\label{fig:env_examples}
\end{figure}

To ensure that our initial model for $D$ does not overfit, we allow our BC-initialized policy to complete a small number of episodes in the environment before we start execution of our method. This tactic is employed in \cite{fujimotoAddressingFunctionApproximation2018, kostrikovDiscriminatorActorCriticAddressingSample2019}, but because our policy is initialized with BC, the actions will not be random and it will be easier for a human to cut an episode short if the agent executes an unsafe action.

Given that our algorithm follows a reinforcement learning paradigm, and that we could easily learn $Q$ in addition to learning $D$ as is done in \cite{fujimotoAddressingFunctionApproximation2018, kostrikovDiscriminatorActorCriticAddressingSample2019}, a natural question is whether we could expand \Method \space to include a self-improving reinforcement learning  component, as is done in \cite{goecksIntegratingBehaviorCloning2020, jenaAugmentingGAILBC2020, rajeswaran*LearningComplexDexterous2018}. In each of these works, it was shown that attempting to learn a policy with an RL framework after initializing with BC either required careful tuning of weighting parameters for combined loss functions, or that initializing the policy with BC actually caused the final performance of the policy to deteriorate considerably \cite{jenaAugmentingGAILBC2020}. On top of that, attempting to learn using RL would require us to add exploration noise to our policy, decreasing the performance and potentially requiring even more expert interactions to learn a satisfactory policy. Due to the varying results in existing literature, we choose to leave this extension as future work.

\section{Experiments and Results}
\begin{figure*}[h!]
   	\includegraphics[width=.98\textwidth, height=4cm]{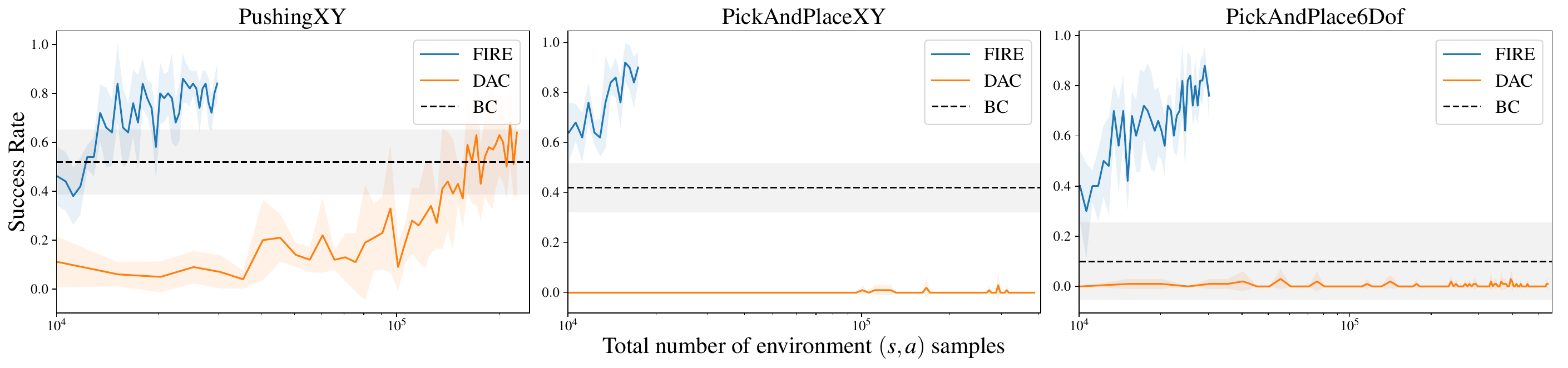}
   	\caption{Performance of \Method \space in our test environments, compared with the total number of environment samples. \Method, Discriminator Actor Critic (DAC) \cite{kostrikovDiscriminatorActorCriticAddressingSample2019}, and BC were all started with the same number of expert trajectories. Of course, in \Method, more expert data was added throughout execution. The success rate of DAC starts off significantly lower than BC or \Method \space because naively initializing RL-based methods with BC has been shown to be detrimental to learning (see end of \cref{sec:other_alg_cons}).}
   	\label{fig:policy_performance}
\end{figure*}

To demonstrate the efficacy of our method, we present experiments where we attempt to learn good policies (measured by success rate) in a variety of challenging manipulation environments (see \cref{fig:env_examples}) described below: 

\begin{enumerate}
	\item \label{it:push_env} \textit{PushingXY}: A pushing environment, in which a narrow, cylindrical end-effector must push a block to be close to another block. The location and rotation of the first block are randomized between trials to positions within a 25 cm x 10 cm rectangle.
	\item \label{it:pick_env} \textit{PickAndPlaceXY}: A pick-and-place environment, in which a two-fingered gripper must grab a large cylinder and place it on a coaster. The location of the cylinder is randomized between trials to positions within a 25 cm x 10 cm rectangle.
	\item \label{it:pick6_env} \textit{PickAndPlace6Dof}: A six degrees-of-freedom pick-and-place environment, where the end-effector can translate and rotate freely, and must grab a long block and place it on a plate. The location of the center of the block is randomized between trials to positions within a 5 cm x 5 cm square, while the rotation is randomized to all possible orientations. The gripper must rotate, often significantly, from its initial position to reach an orientation where it is possible for it to grab the block (see \cref{fig:env_examples}).
\end{enumerate}

Environments \labelcref{it:push_env} and \labelcref{it:pick_env} are constrained to only allow movement in two dimensions. As shown in \cref{fig:policy_performance}, a state-of-the-art IRL method \cite{kostrikovDiscriminatorActorCriticAddressingSample2019} still struggles to solve even these two-dimensional environments adequately, failing altogether in the pick and place environment. The expert controls the agent in Environment \labelcref{it:pick6_env} with an HTC Vive virtual reality hand tracker. Although we do not have access to a dense reward, we do have a binary success signal that determines if, for example, the block is on the plate. The states in each of these environments consist of the poses of the task objects and the end-effector as well as the gripper link positions, while the actions are the end-effector velocity and, for the pick and place environments, a binary signal indicating whether to open or close the gripper. Our environments are simulated in PyBullet \cite{coumans2019} with a UR10 arm and a two-fingered gripper. A timestep in each environment is 0.1 s, while, at the level of simulation, each action is repeated 10 times, as is common in robotics learning environments \cite{plappertMultiGoalReinforcementLearning2018}. Since we are exclusively dealing with the finite-horizon problem, for environments \labelcref{it:push_env}, \labelcref{it:pick_env}, and \labelcref{it:pick6_env}, $T$ is 70, 70, and 80 respectively.

In our experiments, we attempt to answer the following questions: 

\begin{enumerate}
	\item Does \Method \space learn policies that can have strong performance with lower sample complexity than a state-of-the-art IRL-based method and with greater robustness than BC?
	\item How accurately does our system predict failures? Does our failure prediction accuracy improve as more data is added?
	\item How does the failure prediction performance of \Method \space compare to an existing, alternative method?
	\item Is there empirical justification for this method of predicting failures? Do successful episodes tend to result in $D \ge 0.5$, while failures result in $D < 0.5$?
\end{enumerate}

\subsection{Evaluating Policy Performance}

In this section, we evaluate the performance of policies trained using \Method. 
Because our method is inherently based on human decision-making, we validate the efficacy of our method by completing experiments in a variety of environments requiring different manipulation capabilities.
Given the previously described difficulty of generating offline human labels in continuous domains, we do not attempt to benchmark against DAgger. In particular, we expect that it would be nearly impossible to supply human-generated offline expert labels to the six degrees-of-freedom environment. 

We first compare \Method \space with BC alone. Since \Method \space requires the collection of progressively more expert data, a natural question is whether an expert could simply provide more demonstrations initially and achieve the same performance as that found in \Method. To answer this question, in \cref{fig:data_efficiency}, we compare the performance of \Method~with the performance of BC on an equivalent amount of data. The BC data are all collected from full episode rollouts, without any interactive component. The policies learned with data collected from \Method \space clearly show substantial improvement over policies trained with behavioral cloning alone.

To illustrate the sample efficiency in terms of total execution time, we also compare against a state-of-the-art sample-efficient inverse reinforcement learning method \cite{kostrikovDiscriminatorActorCriticAddressingSample2019} in \cref{fig:policy_performance}. It is clear that \Method~learns with considerably greater sample efficiency than \cite{kostrikovDiscriminatorActorCriticAddressingSample2019}, and performs better than BC alone after only a small number of iterations. Our algorithm allows us to learn policies that complete each task with a success rate of over 80\% with 50 minutes or less of wall clock execution time. We stopped execution of \cite{kostrikovDiscriminatorActorCriticAddressingSample2019} after it had converged (in the pushingXY environment) or after it was sufficiently clear that its sample efficiency, were it to converge, would be at least an order-of-magnitude worse than \Method.

\subsection{Evaluating Failure Prediction Performance}
\label{sec:fail_pred_perf}

\begin{figure}[h!]
    \centering
   	\includegraphics[width=.75\columnwidth]{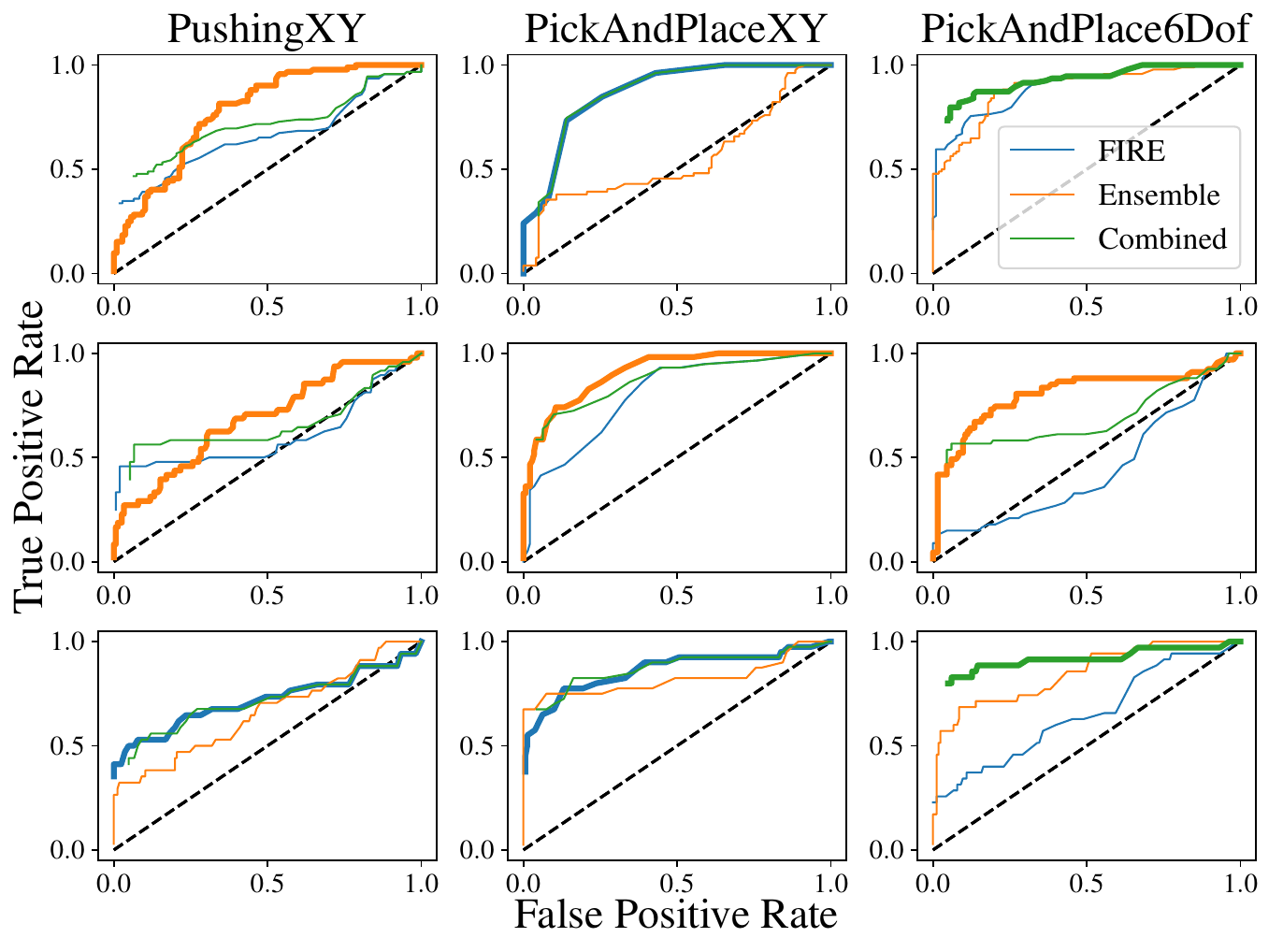}
   	\caption{ROC curves for each environment throughout the execution of \Method, compared with ensemble-based uncertainty \cite{kellyHGDAggerInteractiveImitation2019, mendaEnsembleDAggerBayesianApproach2019} and a method combining both. Columns correspond to environments, while rows correspond to the percentage of time between when we started and when we stopped executing \Method, with the first row being 20\%, the second being 50\%, and the final being 70\%. The method with the largest area under the curve is bolded in each case.}
   	\label{fig:final_rocs}
\end{figure}

\begin{figure}[h!]
   	\includegraphics[width=.98\textwidth]{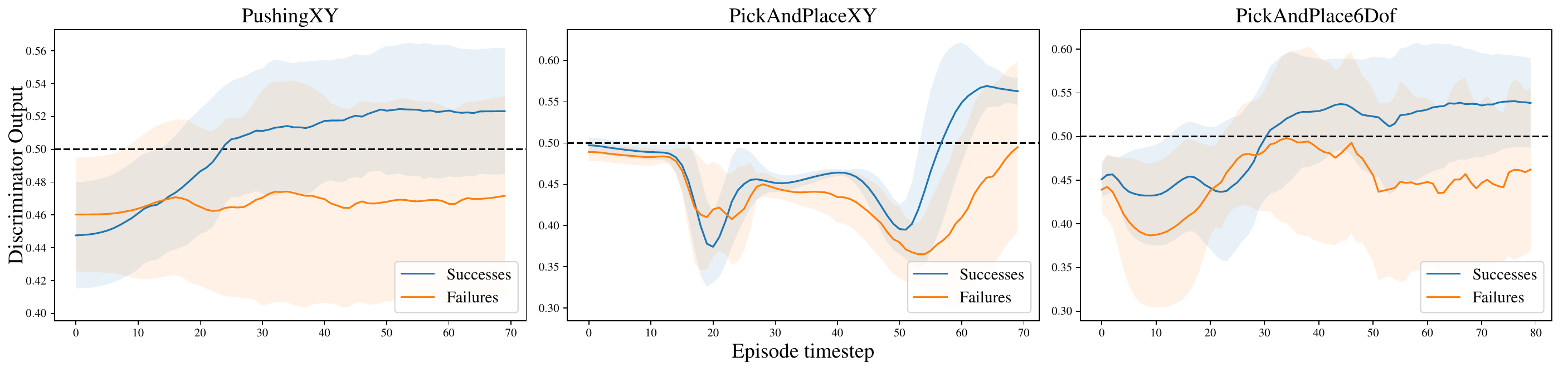}
   	\caption{The average estimate from the discriminator for 200 episodes sampled using a fixed policy and discriminator approximately 70\% of the way through the execution of \Method. The average values for the successful and unsuccessful episodes are separated.}
   	\label{fig:d_avg}
\end{figure}

In addition to showing that \Method \space is able to learn good policies, it is important to know if the failure prediction system is actually capable of predicting failures. To analyze our system's capability in this regard, we consider failure prediction as a binary classification problem, which allows us to use standard statistical measures. We show receiver operating characteristic (ROC) curves (\cref{fig:final_rocs}) for our learned $D$ at several points during the running of \Method.

In existing similar work \cite{kellyHGDAggerInteractiveImitation2019, cuiUncertaintyAwareDataAggregation2019}, the authors have chosen to evaluate their binary predictor based on whether it correctly predicts infractions, which they define as hitting another car or driving off the road. Unfortunately, the concept of an infraction is not so cut-and-dry in manipulation environments. To generate the ROC curves, we used snapshots of our policy weights at various stages of executing \Method~and ran them for 200 episodes, allowing the episodes to fully run to failure or success. From these executions, along with the associated value of $D$ for all steps of each episode, we generated true positive rates and false positive rates for different values of $\beta$, where we define a true positive to be a correct failure prediction.

It is clear from \cref{fig:final_rocs} that in the majority of cases, \Method \space can predict failures at a level above random, though it is stronger in some cases than in others. To answer our original question, it is not clear whether our failure prediction improves as more data is added, as the results are not consistent between environments.

\subsection{Comparing Failure Prediction Performance}

To generate the curves based on the method from \cite{kellyHGDAggerInteractiveImitation2019, mendaEnsembleDAggerBayesianApproach2019}, we trained an ensemble of 10 policies on all of the expert data up to various points during the execution of \Method. Also, we used the $\ell_2$-norm doubt metric $d(s_t) = \Norm{\text{diag}(C_t)}_2$, where $C_t$ is the current covariance matrix of the output of the ensemble. We used the same 200 episode rollouts that we used for evaluating \Method. We consider a prediction to be a true positive whenever $d(s_t)$ is greater than the test value of the parameter at any point during an unsuccessful trajectory, and a true negative when $d(s_t)$ is less than the test value for an entire successful trajectory.

It is clear from \cref{fig:final_rocs} that there are some cases where our method for predicting failures performs best, and some cases where the ensemble-based doubt metric performs best. However, in general, combining the two methods tends to provide at least reasonable performance even if the ensemble metric performs the best, and performs either the best or nearly equivalent to the \Method-only metric in other cases. We acknowledge that our technique for combining the two methods is quite basic: we simply take the logical \text{OR} of the two predictions at every timestep. We expect that a more intelligent combination of the two methods would provide the best performance overall.

\subsection{Further Analysis}

In \cref{fig:d_avg}, we show the average predicted $D$ values for 200 full episodes of each environment using the $D$ and $\pi_{\theta}$ fixed from about 70\% of the way through the execution of \Method. We separate the curves into averages for successful and unsuccessful episodes. By setting $\beta$ to a number of timesteps where $D$ is predicted to be, on average, consistently above 0.5 for successful episodes and consistently below 0.5 for failed episodes, we can predict failures correctly at least some of the time. For example, for the PushingXY, PickAndPlaceXY, and PickAndPlace6Dof environments respectively, we can expect that values of $\beta\ge30$, $\beta\ge35$, and $\beta\ge50$  should predict failures correctly more often than not. However, a clear limitation of this approach is that, should the agent begin to fail significantly before the timestep where it tends to fail most often, the predictor, limited by its higher value of $\beta$, will show a significant lag in predicting the failure. We acknowledge that setting the cutoff for $D$ to 0.5, while intuitive, is somewhat arbitrary. %
In this work, we chose to demonstrate that the values of $D$ \textit{can} be used to predict failures, even with a very simple decision rule. It is also worth noting that a value of $\beta$ that approaches the horizon length $T$ will become less and less useful. Our scheme, which allows the human expert to adjust $\beta$, indirectly ensures that $\beta$ will only approach the horizon length once the policy is succeeding almost all of the time.

\subsection{Parameter Settings}

We use the discriminator gradient-penalty trick from \cite{gulrajaniImprovedTrainingWasserstein2017} to learn a discriminator $D$ that avoids overfitting. We set our network and learning hyperparameters the same as in \cite{kostrikovDiscriminatorActorCriticAddressingSample2019}: a 2-layer MLP policy with 256 hidden units, ReLU activations and a \texttt{tanh} acvitation for the output, and a 2-layer MLP discriminator with 256 hidden units and \texttt{tanh} activations. We train using the Adam optimizer \cite{kingmaAdamMethodStochastic2015} with learning rates of $10^{-3}$ and $10^{-4}$ for the policy during behavioral cloning and episode training respectively. We set our periodic full batch retraining parameter $d$ to 500, and we allow our initially BC-trained policy to collect 10,000 $(s,a)$ pairs (or roughly 125 episodes, 15 minutes of wall clock time) before we start training $D$. When we rerun BC on all of our expert data, we always warm-start with the existing policy since we empirically found this to give better results. Intuitively, we expect that this is because the corrections that the human expert provides specifically address mistakes made by a particular policy, and restarting from scratch wastes this advantage.

In our experiments, we set $\delta_{\text{fn}}$ as 3 and $\delta_{\text{fp}}$ to be 1 because we consider the cost of a false negative to be higher than that of a false positive. A false negative causes the agent to fail or requires the human expert to react quickly themselves, while a false positive merely requires the expert to allow execution to continue, with no requirement for a fast reaction. For each run of \Method, we initialize $\beta$ to 20, and it quickly self-adjusts to the expert's preferences. The pushing and 3D pick-and-place environments are initialized with 50 expert trajectories, while the 2D pick-and-place environment is initialized with 15 expert trajectories.

\section{Conclusion and Future Work}

In this note, we presented \Method, a method for learning a policy from expert demonstrations followed by expert interventions based on failure prediction. Our method predicts failures with a discriminator and an adjustable threshold indicating the tolerable number of non-expert $(s,a)$ pairs in a row. We showed that this technique can be used to learn high-performance policies in several challenging manipulation environments with an order of magnitude better sample efficiency than a state-of-the-art inverse reinforcement learning method. As well, policies learned using our method exhibit significantly better final performance than then those learned using an equivalent amount of data and pure BC. %

To allow policies learned from data and stored in deep neural networks to execute in the real world, we need to better understand when and if such policies are going to fail. Most state-of-the-art methods for failure prediction are based on uncertainty estimated from ensembles or pseudo-ensembles. As we showed in this work, these methods can be partially effective, but are limited by the fact that they do not directly learn from non-expert data.

Future research will formally prove that on-the-fly interventions benefit policy performance, since at the moment, this method is only justified empirically. As well, it is worth exploring in greater depth the differences, empirically and theoretically, between our method for predicting failures and one based on ensemble uncertainties.

\bibliographystyle{IEEEcaps}
\bibliography{refs}

\end{document}

%% file: notation-shortcuts.tex
\usepackage{xparse}

\NewDocumentCommand\bbm{}{ \begin{bmatrix} }
\NewDocumentCommand\ebm{}{ \end{bmatrix} }

\NewDocumentCommand\Norm{m}{\left\Vert#1\right\Vert }

\NewDocumentCommand\Real{}{ \mathbb{R} }

\NewDocumentCommand\DE{}{\mathcal{D}_E}
\NewDocumentCommand\Dpi{}{\mathcal{D}_\pi}
\NewDocumentCommand\pit{}{\pi_\theta}
\NewDocumentCommand\pith{}{\pi_\theta}
\NewDocumentCommand\piE{}{\pi_E}

\NewDocumentCommand\ExpectationSamp{mm}{ \mathbb{E}_{#1}\left[#2\right] }

\NewDocumentCommand\Method{}{FIRE}
\NewDocumentCommand\MethodLong{}{Failure Identification to Reduce Expert burden}